\documentclass{article} % For LaTeX2e
\usepackage{iclr2026_conference,times}

\usepackage{amsmath,amsfonts,bm}

\def\eqref#1{equation~\ref{#1}}
\def\1{\bm{1}}

\DeclareMathAlphabet{\mathsfit}{\encodingdefault}{\sfdefault}{m}{sl}
\SetMathAlphabet{\mathsfit}{bold}{\encodingdefault}{\sfdefault}{bx}{n}

\usepackage{hyperref}
\usepackage{url}
\usepackage{booktabs}
\usepackage{multirow}
\usepackage{graphicx}
\usepackage{amsmath}
\usepackage{amssymb}
\usepackage[T1]{fontenc}
\usepackage[most]{tcolorbox}
\newtcolorbox{promptbox}[1][]{%
  breakable, enhanced, sharp corners,
  colback=gray!4, colframe=gray!60, boxrule=0.4pt,
  left=2.2mm, right=2.2mm, top=1.4mm, bottom=1.4mm,
  fonttitle=\bfseries\sffamily\footnotesize, coltitle=black,
  colbacktitle=gray!18, titlerule=0pt,
  fontupper=\footnotesize\ttfamily,
  title={#1}}

\title{SportD: How do VLMs physically strategize?\\}

\author{%
\hspace*{-0.6cm}% nudge the author block slightly left of centre
\begin{tabular}{c}
Jasin Cekinmez$^{*1}$ \quad Addison J. Wu$^{*1}$ \quad Haotian Xia$^{2}$ \quad Kyumin Andrew Shim$^{3}$ \quad Anay Putty$^{3}$ \\[2pt]
Jinglin Xiao$^{4}$ \quad Zhuohan Liu$^{3}$ \quad Leo Liu$^{5}$ \quad Weining Shen$^{3}$ \\[6pt]
{\small \textmd{$^{1}$ Princeton University} \quad \textmd{$^{2}$ Rice University} \quad \textmd{$^{3}$ UC Irvine} \quad \textmd{$^{4}$ NYU} \quad \textmd{$^{5}$ UC Santa Barbara}} \\[4pt]
{\small \textmd{$^{*}$ Equal contribution}} \\[4pt]
{\small \textmd{\texttt{\{jasincekinmez, addisonwu\}@princeton.edu}}}
\end{tabular}%
}

\newcommand{\benchname}{\textbf{SportD}}

\iclrfinalcopy  % preprint/arXiv mode: reveal authors, drop line numbers (revert to comment for blind submission)

\begin{document}

\maketitle
% preprint header: blank the "Published as a conference paper..." banner and its rule
\fancyhead{}
\renewcommand{\headrulewidth}{0pt}

\begin{abstract}
Vision--language models (VLMs) can describe a scene, but can they \emph{act} well within one? We study whether VLMs can make sound strategic decisions, using soccer as an objective testbed with quantifiably-valued actions. We introduce \textbf{\benchname{}}, a dataset and evaluation consisting of 1421 decision scenarios across professional men's and women's soccer games, where a VLM must decide what action to take next. Models on average select the optimal action around 27\% of the time, less often than the professional players, and capture markedly less of the value at stake. Furthermore, they exhibit a clear preference for safer actions, favoring lower-variance, lower-value choices that also make less physical progress toward goal. Frontier VLMs are better at estimating whether an action will succeed, placing the highest-success-probability action among their top choices in 72--85\% of cases. Yet VLMs systematically conflate likelihood with value, assigning higher value to actions that are more likely to succeed ($\rho=+0.30$ to $+0.52$), despite no such relationship in the ground truth ($\rho=-0.08$). Modifying the deliberation instructions to encourage risk-taking brings the frontier models closer to the players’ skill levels. \benchname{} opens a new direction for rigorously evaluating physical strategic decision-making in VLMs, showing that careful decomposition of their choices can reveal the mechanisms underlying systematic biases such as risk aversion. The dataset and code are available at \url{https://github.com/JasinCekinmez/SportD}.

\end{abstract}

\section{Introduction}
Vision-language models (VLMs) have become strong at understanding visual scenes, but whether they can reason strategically about how to act in the real physical world remains unclear. This requires models to integrate fine-grained spatial perception with reasoning over possible actions and their consequences. Such reasoning is essential for agents to operate in the physical world, but it is largely absent from benchmarks centered on recognition, captioning, and visual question answering, which evaluate representations of a scene rather than decisions within it.

Team sports let us study a broader question: can VLMs turn visual understanding into good strategic actions? Existing sports benchmarks primarily test understanding \citep{sportu2025, sportr2026} of what is happening, while  benchmarks evaluating strategic-reasoning capabilities largely operate in abstract or simulated environments \citep{vsbench2026}. We instead test strategic decision-making in the physical world, asking VLMs to select the next action given an observation of the environment, and evaluating their choices against actions that can be quantifiably valued.

\begin{figure}[t]
\centering
\includegraphics[width=\textwidth]{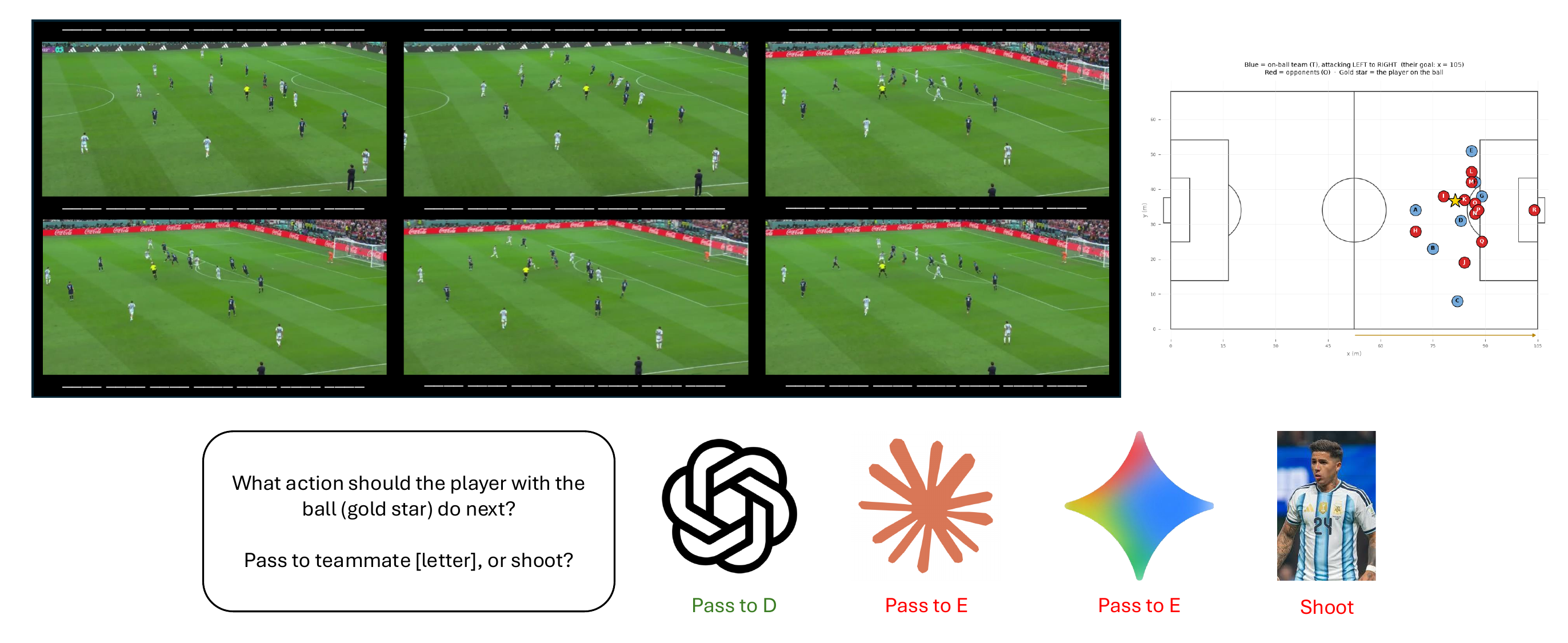}
\caption{\textbf{The \benchname{} task.} From the run-up clip, decision frame, and top-down map (lettered circles for players, a gold star for the player on the ball, an
arrow for the attacking direction), the VLM chooses one action: a pass to a lettered teammate or a
shot. \emph{Bottom right:} the three VLMs' and the real player's choices, green if optimal and red
otherwise. Here one VLM, GPT 5.6 Sol, finds the optimal pass; the others, and the player (who shot), do not.}
\label{fig:core}
\end{figure}

We introduce \textbf{\benchname{}}, a comprehensive evaluation and dataset of 1421 on-ball situations from the 2022 men's and 2023 women's World Cups (Fig.~\ref{fig:core}). Soccer provides a useful setting for studying strategic physical decision-making: each possession unfolds among many moving players,  admitting many plausible actions. Soccer also offers rich event and tracking data, together with established models for valuing actions. For each example, a VLM observes the seconds leading up to an on-ball situation and selects the next action, either passing to a specific teammate or shooting. Because every candidate action can be assigned a quantitative value, \benchname{} measures not only whether a model selects the best action, but whether its suboptimal choices follow systematic patterns.

Our findings shed light on the limitations of VLMs in strategically reasoning about the physical world. In \benchname{}, VLMs select the optimal action on 27.4\% of decisions on average, less often than the professional players (35.1\%), and capture far less of the value at stake (mean skill 0.18 versus 0.34). More importantly, VLMs are risk averse, with models choosing lower-variance actions than the optimum. The models' internal beliefs reveal a mismatch between likelihood and value, identifying which actions are most likely to succeed with 72--85\% recall, but the most valuable actions only 50--58\% of the time. Moreover, while success probability and expected value are not correlated in the value model ($\rho=-0.08$), every VLM assigns them a positive relationship ($\rho=+0.30$ to $+0.52$). These results suggest that VLM risk aversion arises not merely from poor perception or imitation, but from systematically treating what is easier to execute as what is more valuable.

\section{\benchname{}}
\label{sec:task}

\subsection{Dataset}

\benchname{} contains 1421 on-ball decisions sourced from two professional tournaments --- 726 from the 2022 men's World Cup and 695 from the 2023 women's World Cup \citep{statsbomb}. Each decision $d$ is a single moment at which the player on the ball must act, and we present to the VLM (i) images from the previous $\sim\!5$\ seconds, sampled at 0.5 second intervals; (ii) an image of the moment immediately preceding the decision; and (iii) a bird's-eye view map showing the positions of all visible players. The main prompt is given in App.~\ref{app:prompt-decision}.

The VLM selects exactly one action $a$ from the available action set $\mathcal{A}_d$, where \begin{equation} \mathcal{A}_d = \{\texttt{SHOOT}\} \cup \{\text{pass to teammate } t : t \in T_d\}. \label{eq:actions} \end{equation}
Here, $T_d$ is the set of visible teammates in the image. The median event offers approximately six candidate actions.

\subsection{Value model}
We value actions with the VAEP (Valuing Actions by Estimating Probabilities) model \citep{decroos2019actions}. Each match is first converted into SPADL (Soccer Player Action Description Language), a unified representation of on-ball actions in which each action is described by its type, start and end location, body part, and outcome. VAEP values an action by how it changes the acting team's chances of scoring and conceding. From each game state, separate models estimate the probabilities that the team scores or concedes within the next ten actions, giving the state value $
V = P_{\text{score}} - P_{\text{concede}}.$

The value of an action is the change in $V$ from immediately before to immediately after it. We fit separate value stacks for the men's and women's benchmarks. The men's stack is trained on Euro 2020 and Euro 2024, with the 2022 men's World Cup reserved for evaluation. The women's stack uses the same specification, trained on the Euro 2022 and Euro 2025 women's matches, and evaluated on the 2023 women's World Cup.

\subsection{Metrics}
\label{sec:vaep}
\label{sec:metrics}
\subsubsection{Action valuation}
We assign a numerical value to every action with VAEP \citep{decroos2019actions}, which scores an action by how much it changes the team's chance of scoring next. Higher-valued actions are deemed more optimal to carry out. An action succeeds with some probability and yields
one value if it does and another if it does not, so its expected value is
\begin{equation}
V(a) \;=\; p_a\, v^{+}_a + (1 - p_a)\, v^{-}_a ,
\label{eq:ev}
\end{equation}
where $v^{+}_a$ and $v^{-}_a$ are the values of the game state that results if the action succeeds or fails, respectively, and $p_a$ is its success probability. For a pass, $p_a$ is the completion probability and $v^{+}_a$, $v^{-}_a$ are the values of keeping possession at the target location versus losing it. For a shot, $p_a$ is the expected-goal probability \citep{lucey2015quality} with a scored goal worth $v^{+}_a=1$. 

\subsubsection{Scoring a decision}

\paragraph{Regret}
Regret is the value lost relative to the optimal action. For a chosen action with value $V$,
\begin{equation}
R = V^\star - V \ge 0 ,
\label{eq:regret}
\end{equation}
where $V^\star$ is the value of the best available action. If the optimal action was selected, then $R$ is 0. Better actions have lower regret $R$.

We group the events into three tiers of human decision quality. The \texttt{low} tier contains the 499 events where the player chose the optimal action ($a^\star$), matching the players' accuracy in Table~\ref{tab:main}. Among the events where the player chose a suboptimal action, the \texttt{medium} tier contains 482 with small regret ($R\le0.034$) and the \texttt{high} tier contains 440 with larger regret ($R>0.034$).

\paragraph{Skill}
Values vary substantially across events, so mean regret can be dominated by a small number of high-value decisions. We report \emph{skill} that rescales each event to its own range,
\begin{equation}
s \;=\; \frac{V(\text{chosen action}) - \bar V}{V^\star - \bar V} ,
\label{eq:skill}
\end{equation}
where $\bar V$ is the average value of the actions on offer. Skill is $1$ for the best action, $0$
for a choice no better than picking at random, and negative for worse.

\paragraph{Risk}
To measure whether a VLM prefers \emph{safe} actions, we use the variance of each action's payoff under the same value model,
\begin{equation}
\textrm{Risk}(a) = p_a(1-p_a)\big(v_a^{+}-v_a^{-}\big)^2 ,
\label{eq:var}
\end{equation}
where $p_a$ is the probability that action $a$ succeeds, and $v_a^{+}$ and $v_a^{-}$ are its values if it succeeds or fails. 

\section{Results}
\label{sec:exp}

\subsection{VLMs perform worse than professional players}
\label{sec:exp-main}
We evaluate five frontier VLMs, comprising of GPT 5.6 Sol, Claude Opus 4.8, Gemini 3.5 Flash, Qwen3-VL-8B, and Llama 4 Maverick on \benchname{}. As shown in Table \ref{tab:main}, no VLM matches the players' accuracy of 35.1\% or mean skill of 0.34. On average, proprietary VLMs attain an accuracy of 30.2\% and mean skill of 0.24, and the open-source models, Qwen3-VL-8B and Llama-4-Maverick, attain the lowest accuracy and skill, at roughly 23\% accuracy with skills 0.07 and 0.11, respectively.

\begin{table}[t!]
\centering
\caption{Main results. Accuracy (absolute Top-$X\%$), skill, and risk percentile. Top-$X\%$ is the fraction of choices ranking among the best $X\%$ of available actions.\\}
\label{tab:main}
\begin{tabular}{lccccc}
\toprule
strategy & accuracy & top-20\% & top-30\% & mean skill & risk percentile \\
\midrule
real player & 35.1\% & 49.8\% & 57.4\% & 0.335 & 60 \\
\midrule
GPT 5.6 Sol & 34.6\% & 46.3\% & 54.1\% & 0.298 & 50 \\
Gemini 3.5 Flash & 30.3\% & 43.3\% & 51.6\% & 0.263 & 53 \\
Claude Opus 4.8 & 25.8\% & 36.7\% & 44.3\% & 0.157 & 46 \\
\midrule
Qwen3-VL-8B & 23.1\% & 33.5\% & 40.3\% & 0.069 & 53 \\
Llama-4-Maverick & 23.2\% & 34.9\% & 41.3\% & 0.107 & 48 \\
\bottomrule
\end{tabular}
\end{table}

\begin{table}[t!]
\centering
\caption{Per-tournament breakdown of the headline metrics shown in Table~\ref{tab:main}.}
\label{tab:women}
\begin{tabular}{lcccccc}
\toprule
 & \multicolumn{2}{c}{accuracy} & \multicolumn{2}{c}{mean skill} & \multicolumn{2}{c}{risk percentile} \\
strategy & men & women & men & women & men & women \\
\midrule
real player & 35.3\% & 35.0\% & 0.315 & 0.356 & 60 & 60 \\
\midrule
GPT 5.6 Sol & 34.0\% & 35.1\% & 0.268 & 0.330 & 51 & 48 \\
Gemini 3.5 Flash & 29.1\% & 31.6\% & 0.224 & 0.304 & 56 & 51 \\
Claude Opus 4.8 & 20.8\% & 31.1\% & 0.085 & 0.233 & 46 & 46 \\
\midrule
Qwen3-VL-8B & 23.7\% & 22.4\% & 0.070 & 0.069 & 54 & 53 \\
Llama-4-Maverick & 23.3\% & 23.2\% & 0.091 & 0.123 & 49 & 47 \\
\bottomrule
\end{tabular}
\end{table}

\begin{figure}[t!]
\centering
\includegraphics[width=0.62\textwidth]{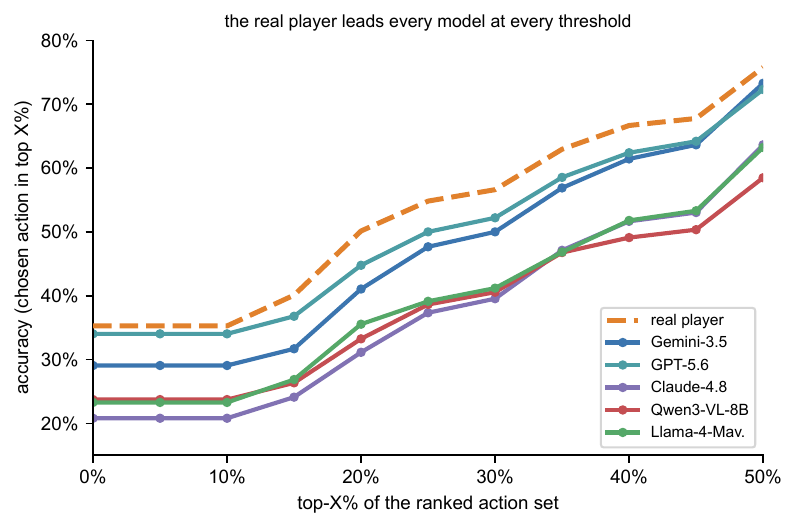}
\caption{Rank-percentile accuracy: fraction of events on which the chosen action falls within the
top $X\%$ of the ranked action set. The real player attains a higher adjusted accuracy than every VLM at every threshold. The comparison is purely ordinal.}
\label{fig:rankcurve}
\end{figure}

Even if we consider an answer to be correct if it ranks among the best $X\%$ of the available actions, models are still worse than humans. At a top-20\% threshold, the VLMs reach 33.4--46.3\% versus 49.8\% for the real player, and the ordering holds at every threshold of the full rank-percentile curve (Fig.~\ref{fig:rankcurve}). Relaxing the metric does not close the performance gap between the VLMs and the player. Even when we make the visual format easier for the VLM to parse by labeling the players on the actual gameplay scene directly, there is no meaningful difference in accuracy (Appendix \ref{app:labeled}). 

\begin{figure}[h]
\centering
\includegraphics[width=0.5\textwidth]{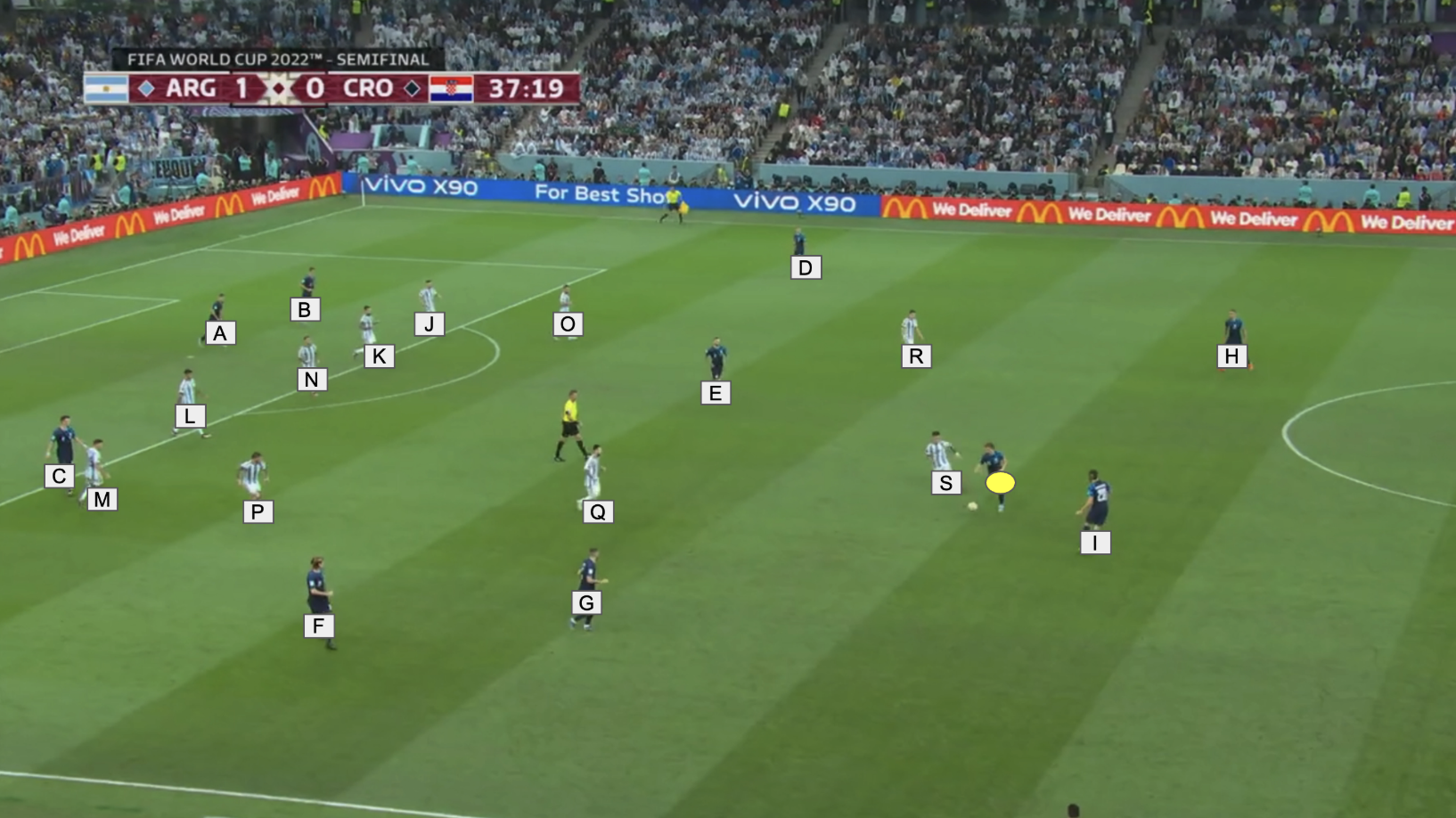}
\caption{Example of an image with labels for all players and the on-ball player (yellow oval) used for the easier visual control experiment (Appendix \ref{app:labeled})}
\label{fig:labeled}
\end{figure}

\subsection{VLMs carry out risk-averse actions}
\label{sec:exp-risk}

\begin{figure}[t]
\centering
\includegraphics[width=\textwidth]{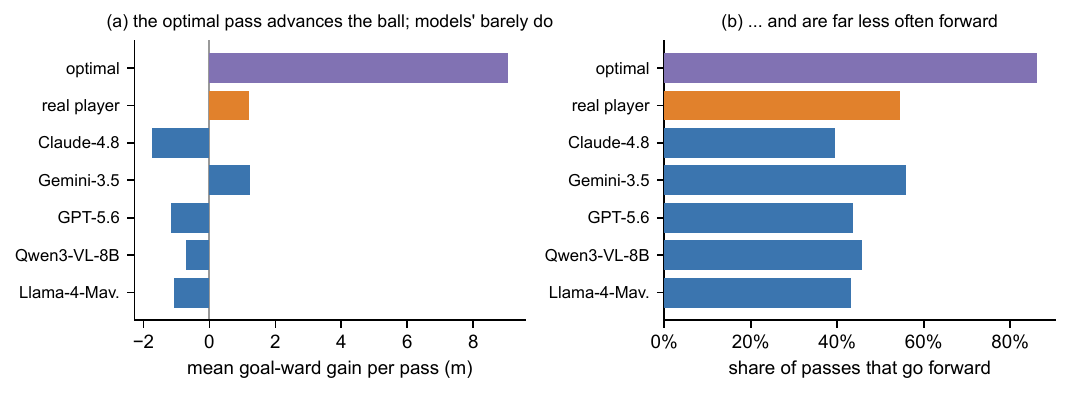}
\caption{Progression of the chosen pass. (a) Mean goal-ward gain (meters the ball moves toward the
attacking goal): the optimal pass advances it $\sim$$9$\,m, the real players $\sim$$2$\,m, and the
VLMs near zero or backward. (b) Share of passes that go forward at all.}
\label{fig:progression}
\end{figure}

The models' conservatism is visible directly in the geometry of their passes. For each pass selected by the VLM, we measure \emph{goal-ward gain}: the number of meters the ball moves toward the attacking goal (Fig.~\ref{fig:progression}). The optimal pass is decisively forward, advancing the ball $9.0$\,m on average and progressing toward goal on $86\%$ of events. The VLMs' passes barely move the ball forward --- Gemini $+1.2$\,m, GPT $-1.2$\,m, Claude $-1.7$\,m (i.e.\ near zero or backward) --- going forward on only $40$--$56\%$ of events, comparable to the real players ($+1.2$\,m, $55\%$) yet far short of the optimum, even though the magnitudes of the pass distances between VLMs and the optimal baseline are nearly equal ($14$--$15$\,m vs.\ $15$\,m for $a^\star$). 

We see similar trends when we look at the risk metric defined in \ref{sec:metrics}. Across all possible decision points, the optimal action lies at the 75th risk percentile and the real player's choice at the 60th. VLMs select actions much closer to the middle of the distribution (46th--53rd percentile; Table~\ref{tab:main}). This conservatism carries a clear value cost: mean skill ranges from 0.067--0.298 across VLMs, compared with 0.335 for the real player. The models therefore systematically favor safer actions while leaving value unrealized.

\begin{figure}[t]
\centering
\includegraphics[width=0.62\textwidth]{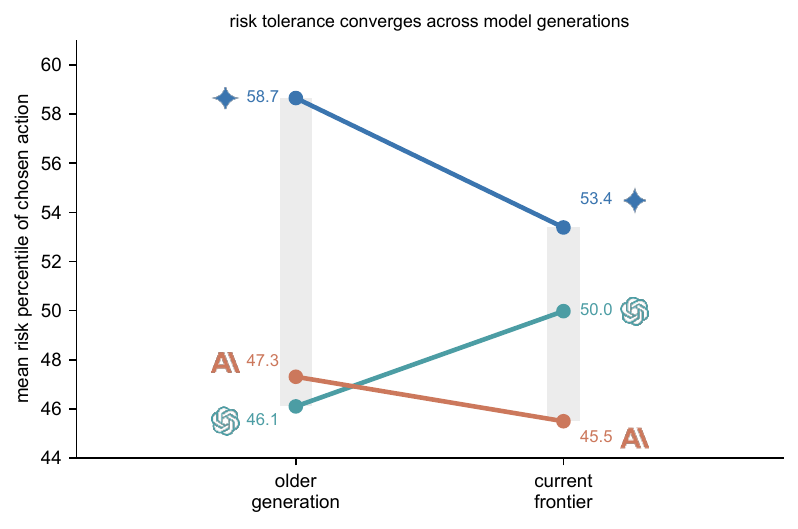}
\caption{Risk tolerance converges across model generations. The $y$-axis is the mean within-event risk percentile of the chosen action. Each line is one model family, from its older generation to the current frontier. The spread shrinks from $10.8$ to $7.9$ points. The families converge from both sides: GPT rises ($45.4 \to 50.0$), Gemini falls ($56.2 \to 53.4$), and Claude edges down ($48.1 \to 45.5$).}
\label{fig:riskconverge-pctile}
\end{figure}

Risk tolerance converges across model generations. We run the same evaluation on older model generations: GPT-4o, Gemini 2.5 Flash, and Claude Haiku 4.5. The older models are more divergent in their risk profiles, with their mean risk percentile of the chosen action scattering across $10.8$ points. Gemini 2.5 Flash is the most aggressive at $56.2$, GPT-4o the most conservative at $45.4$, and Claude Haiku 4.5 between them at $48.1$. The current frontier roster of models collapses into a $7.9$-point band ($45.5$--$53.4$; Fig.~\ref{fig:riskconverge-pctile}). The families converge from both sides: GPT rises, Gemini falls, and Claude edges down. Increasing capability did not diversify risk tolerance across the frontier, instead \emph{standardizing} it.

\subsection{The models estimate likelihood of success well, but value poorly}
\label{sec:exp-belief}

Why are VLMs risk-averse? One possibility is a distortion in their internal consistency in which they may misjudge the consequences and returns of actions that are more likely to succeed. But do they even know which actions are more likely to succeed in the first place? In human decision-making, probability and value are separate inputs to choice, yet people conflate them: a single impression can drive both risk and benefit judgments \citep{finucane2000affect}, so that safer-seeming actions are also judged more valuable. We test for the same conflation by eliciting each model's beliefs on the same events as two separate rankings (App.~\ref{app:prompt-belief}): one orders every legal action by its probability of success, the other by its expected value. Each is prompted in a fresh context so that neither response can condition on the other.

\begin{figure}[t]
\centering
\includegraphics[width=\textwidth]{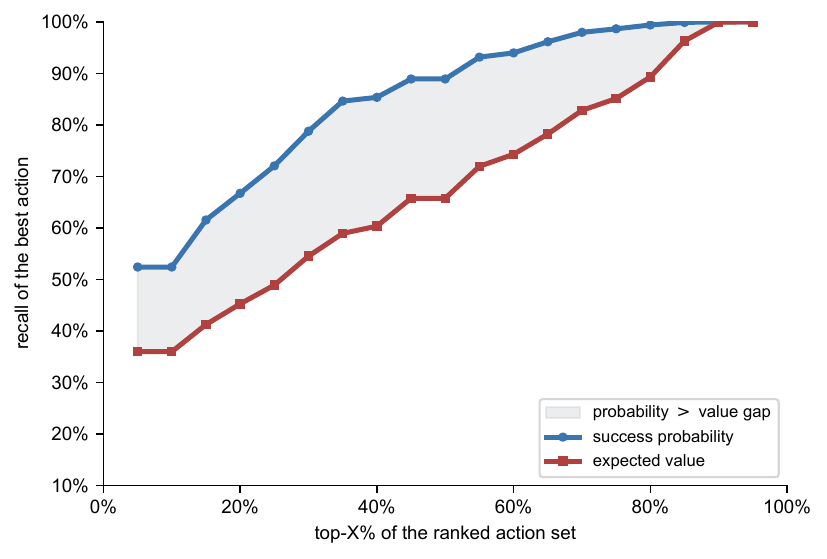}
\caption{Recall of the best action as the top-X\% cutoff sweeps. Success-probability recall (blue) exceeds expected-value recall (red) at every cutoff; the gap is shaded. The models rank likelihood far better than value.}
\label{fig:belief-sweep}
\end{figure}

We evaluate each model's ranking with recall at the top third of the ranked actions: whether the truly best action appears among the top $33\%$. Figure~\ref{fig:belief-sweep} sweeps every cutoff. Models identify which actions are most likely to succeed well, reaching success-probability recall of $72$--$85\%$. They identify which actions are most valuable far less well, with expected-value recall of only $50$--$58\%$. The gap holds at every cutoff (Fig.~\ref{fig:belief-sweep}): the models rank safety far more reliably than value. Table~\ref{tab:belief-recall} gives the per-model recall.

\begin{figure}[t]
\centering
\includegraphics[width=\textwidth]{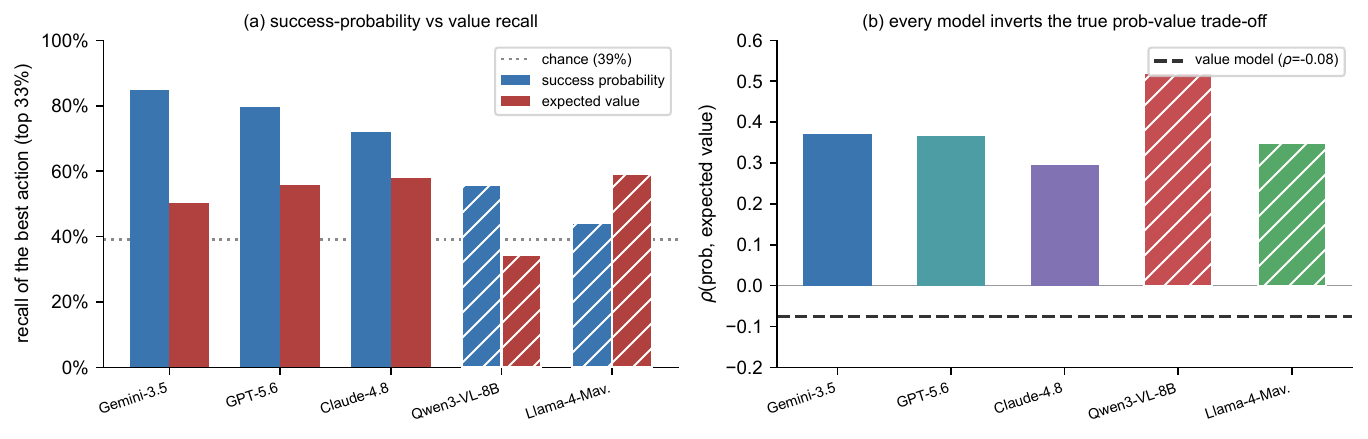}
\caption{VLMs estimate likelihood better than value.
Most models identify high-success-probability actions far more reliably than high-value actions (left), and systematically assign higher value to actions they believe are more likely to succeed (right), reversing the ground-truth probability--value relationship.}
\label{fig:belief}
\end{figure}

These value errors follow a clear pattern. Models systematically overvalue actions that are more likely to succeed. Across the available actions an action's success probability and its expected value are
exhibit almost no correlation (Spearman $\rho=-0.08$). Instead, every VLM asserts a
\emph{positive} correlation between probability and expected value ($\rho=+0.30$ to $+0.52$;
Fig.~\ref{fig:belief}b, per-model values in Table~\ref{tab:belief-corr}). The models behave as though the easier action were the better one. The conservatism therefore reflects a distorted notion of value, with models treating success probability as a proxy for how worthwhile an action is.

\subsection{VLMs recover value when steered toward risk, and forfeit it when steered away}
\label{sec:exp-steer}

Motivated by the risk aversion and valuation miscalibration observed in Sec.~\ref{sec:exp-risk} and Sec.~\ref{sec:exp-belief}, we test whether encouraging risk-taking improves VLM decisions. We modify only the deliberation instructions to encourage riskier attacking actions (\textsc{seek}) or prioritize retaining possession (\textsc{avoid}). App.~\ref{app:prompt-steer} provides the full prompts. We evaluate each event on the men's set, using the unmodified benchmark run from Sec.~\ref{sec:exp-main} as the \textsc{neutral} baseline.

\begin{figure}[t]
\centering
\includegraphics[width=0.85\textwidth]{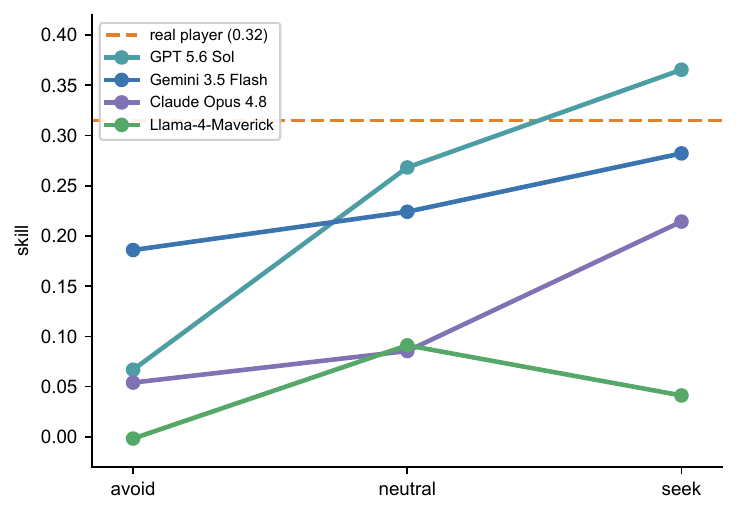}
\caption{Risk-steering the decision prompt on the men's set. Each line is one model; the
dashed line is the real players' skill. Steering from \textsc{avoid} to \textsc{neutral} to
\textsc{seek} raises skill for the closed-source models and lifts GPT to the players' level.
\textsc{avoid} lowers every model, and the weakest model, Llama-4-Maverick, moves the wrong way.}
\label{fig:steer}
\end{figure}

The frontier models are steerable, and risk-seeking prompts recover value. The \textsc{seek} instructions move GPT, Claude, and Gemini toward riskier actions and increase their skill (Fig.~\ref{fig:steer}). GPT rises from $0.27$ to $0.37$, surpassing the real players ($0.32$ on the men's set), while Claude more than doubles from $0.09$ to $0.21$. GPT is the only model to reach the players' level; under \textsc{seek} the other closed models improve but still trail the humans (Gemini $0.28$, Claude $0.21$), and the open Llama does not benefit ($0.04$). \textsc{avoid} moves all models toward safer actions and lowers skill, confirming that their default policies lie on the conservative side of optimal.

\subsection{VLMs match the human actions less as the humans perform worse}
\label{sec:exp-lean}
Do VLMs just copy the human's choice? As the humans' actions grow more suboptimal (larger regret $R$), the frontier models copy it less. Across the three regret strata as established in Section \ref{sec:metrics}, the human-match rate falls by 23, 20, and 15 percentage points for GPT, Gemini, and Claude per $0.1$ increase in $R$, whereas open models already have a low accuracy to begin with so the decline is not as visible there. The models therefore do not simply reproduce human mistakes; they increasingly diverge from the human choice as it becomes more suboptimal.

\begin{table}[t]
\centering
\caption{As the human's action grows more suboptimal (\texttt{low}$\to$\texttt{medium}$\to$\texttt{high})
, the frontier models match the human action less as well.\\}
\label{tab:lean-tier}
\begin{tabular}{lccc}
\toprule
 & \texttt{low} & \texttt{medium} & \texttt{high} \\
 & $=a^h$ & $=a^h$ & $=a^h$ \\
\midrule
GPT 5.6 Sol & 58.7\% & 37.8\% & 31.1\% \\
Gemini 3.5 Flash & 47.1\% & 30.3\% & 23.2\% \\
Claude Opus 4.8 & 35.0\% & 23.2\% & 16.9\% \\
\midrule
Qwen3-VL-8B & 27.5\% & 17.4\% & 19.6\% \\
Llama-4-Maverick & 26.7\% & 28.2\% & 20.2\% \\
\bottomrule
\end{tabular}
\end{table}

\section{Related Work}

\paragraph{Vision--language models and visual reasoning.}
A large body of work evaluates whether VLMs can understand and reason about visual observations, including compositional and relational reasoning \citep{johnson2017clevr,antol2015vqa,hudson2019gqa}, spatial reasoning \citep{liu2023vsr,chen2024spatialvlm}, expert multimodal knowledge \citep{yue2024mmmu}, challenging perceptual failures \citep{fu2024blink,tong2024eyes}, and temporal understanding in video and egocentric settings \citep{xiao2021nextqa,mangalam2023egoschema,li2024mvbench,grauman2022ego4d}. These benchmarks largely evaluate understanding or reasoning \emph{about} visual observations. \benchname{} instead asks whether a VLM can use those observations to choose a strategically valuable action, evaluating not what the model can describe about a scene but what it decides to do within one.

\paragraph{Physical and strategic reasoning.}
Physical-reasoning benchmarks study whether models can predict how scenes will evolve \citep{battaglia2013simulation,bakhtin2019phyre,yi2020clevrer,bear2021physion,piloto2022intuitive}, while world models learn latent dynamics that can support prediction and planning \citep{ha2018worldmodels,hafner2023dreamerv3,bruce2024genie,lecun2022path}. Separately, strategic agents have reached strong performance in games such as Go, chess, poker, Stratego, and real-time strategy \citep{silver2016alphago,silver2018alphazero,brown2019pluribus,perolat2022stratego,vinyals2019alphastar,berner2019dota}, and language-conditioned agents can negotiate and cooperate \citep{cicero2022,park2023generative}. These systems generally act from symbolic or simulated state; similarly, VS-Bench \citep{vsbench2026} evaluates strategic VLM agents in abstract multi-agent environments. \benchname{} connects physical perception and strategic choice by requiring models to select among alternative actions directly from real-world visual observations.

\paragraph{Sports understanding and action valuation.}
Sports analytics provides quantitative models for evaluating decisions, including possession-value and space-control approaches \citep{decroos2019actions,fernandez2021framework,cervone2016epv,fernandez2018wide,spearman2018beyond,lucey2015quality,decroos2020player}. In parallel, multimodal sports benchmarks study action spotting, broadcast understanding, perception, and comprehension \citep{giancola2018soccernet,deliege2021soccernetv2,sportu2025,sportr2026}. To our knowledge, these lines have not been combined to evaluate whether a VLM can make a strategically valuable decision given the current state of the environment it is in, while considering alternatives holistically. Related work on behavioral cloning learns to reproduce expert behavior \citep{pomerleau1991alvinn,ross2011dagger,ho2016gail}, while vision--language--action and generalist agents extend learned action selection to embodied settings \citep{reed2022gato,driess2023palme,driess2023rt2}. \benchname{} instead connects visual understanding with value-based action selection, enabling us to study not only whether VLMs make good decisions, but also the systematic biases underlying the actions they choose.

\section{Conclusion}

We introduce \benchname{}, a benchmark for evaluating whether VLMs can turn visual understanding into strategically valuable actions in the physical world. Rather than asking models to describe what is happening in a scene, \benchname{} asks what a player should do next and evaluates that decision against the estimated value of every available alternative. Across 1421 decisions from the 2022 men's and 2023 women's FIFA World Cups, every VLM we evaluate falls short of the professional players who faced the same situations. More importantly, their errors exhibit a consistent structure.

VLMs systematically favor safer actions. They choose lower-risk actions than the value-optimal alternative on most events and make substantially less progressive passes, sacrificing expected value alongside risk. The models' elicited beliefs point to a miscalibration of valuation. Frontier VLMs are substantially better at identifying which actions are likely to succeed than which actions are most valuable. Yet across all models, estimated success probability is positively associated with estimated value, even though the corresponding relationship under the value model is slightly negative. VLMs therefore behave as though actions that are easier to execute are also more worthwhile. The evidence is therefore most consistent with a systematic distortion in how VLMs translate beliefs about possible outcomes into strategic value, rather than with a deliberate risk--reward tradeoff. Steering models towards favoring more riskier actions increases their average skill, but still lagging behind professional players.

Soccer provides a controlled setting for exposing this distinction because it combines complex real-world visual observations, multiple plausible actions, and quantitative estimates of their consequences. Future work can extend \benchname{} to other sports and physical domains, richer spatiotemporal value models, and broader action spaces including other movements, like dribbling. More generally, our results highlight a capability that descriptive vision benchmarks can miss: a model may understand what is likely to happen without understanding which possible future is worth pursuing. Evaluating this gap will be important as multimodal models are increasingly expected not only to perceive the physical world, but to make decisions within it.

% \subsubsection*{Reproducibility Statement}
% All numbers are computed from per-event JSON records through a single scoring module
% (\texttt{metrics.py}); the analysis notebook regenerates every table and figure from disk with
% no API calls.

\subsection*{AI use statement}
In this work, we used generative AI tools to propose and refine hypotheses, to design and give
feedback on our research methodology and experiments, to implement methods (writing and debugging
the evaluation pipeline, the analysis code, and the figure-generation scripts), to clean and
reformat the dataset (joining and normalizing the answer keys and assembling the released data). We did not use generative AI tools to develop the paper's
conceptual framework or to interpret results. Generating synthetic datasets, formulating mathematical claims, providing critical
ingredients for or writing proofs, translation, and qualitative or thematic data analysis are not
applicable to this work. We additionally used generative AI tools to assist with drafting and
editing the manuscript and to help identify related literature. We have reviewed all AI-assisted
work: every AI-written or AI-modified piece of code was read and tested by the authors, every
reported number was recomputed and checked against the paper, and all AI-assisted text was verified
for accuracy. We take responsibility for the final content of this work, including text, claims, or
artifacts produced with the aid of generative AI.

\bibliographystyle{iclr2027_conference}
\bibliography{references}

\newpage
\appendix

\section{Prompts}
\label{app:prompt}

\subsection{Decision prompt}
\label{app:prompt-decision}
The main benchmark task (Sec.~\ref{sec:task}). The run-up is presented as a sequence of still
frames sampled from the seconds before the decision, one every $0.5$\,s; \texttt{\{n\}} is the
number of run-up frames, \texttt{\{attack\}} the attacking direction, and \texttt{\{teammates\}}
the letters that can legally be passed to.

\begin{promptbox}[Decision prompt (full stimulus: run-up frames + freeze-frame + map)]
You are an elite football (soccer) analyst. Watch this moment and decide what the player on the
ball should do next. Judge it on what is visible at the moment the ball is played, not on hindsight.
\medskip

You are shown one moment from a match, three ways:
\medskip

1. RUN-UP -- \{n\} still frames from the seconds leading up to the moment, one every
\{interval\} seconds of real time, in chronological order. Read them as a sequence: they show who
is moving where, and how fast.\\
2. FINAL FRAME -- the broadcast frame at the exact instant the decision is made. This is the
decision point.\\
3. PLAYER MAP -- a top-down view of that same instant. Every player on camera is a lettered circle.
Blue = the on-ball team (T). Red = the opponents (O). The gold star is the player on the ball. The
arrow shows which way the on-ball team is attacking.
\medskip

THE DECISION\\
The player on the ball is the gold star. \{attack\}
\medskip

What should he do RIGHT NOW? Your options are:\\
~~-- PASS to one of his lettered teammates (the blue circles on the map). The teammates you can
pass to are: \{teammates\}.\\
~~-- SHOOT.
\medskip

Reason it through first: read the shape of the defence, who is actually free, who is covered or
offside, where the space is. Ask whether a shot from here beats the best available pass. Say
concretely why your choice beats the obvious alternatives.
\medskip

Then end your reply with this line and nothing after it:
\medskip

ANSWER: <letter>
\medskip

where <letter> is one teammate letter (e.g. ANSWER: E) or the word SHOOT.
\end{promptbox}

\subsection{Labeled broadcast-frame prompt}
\label{app:prompt-labeled}
The ablation of Appendix~\ref{app:labeled}. The top-down player map is removed; instead the model
sees the run-up frames and the real broadcast frame with every player tagged by a capital letter and
the ball-carrier marked. \texttt{\{n\}} is the number of run-up frames and \texttt{\{teammates\}}
the letters that can legally be passed to.

\begin{promptbox}[Labeled broadcast-frame prompt (run-up frames + labeled final frame, no map)]
You are an elite football (soccer) analyst. Watch this moment and decide what the player on the ball
should do next. Judge it on what is visible at the moment the ball is played, not on hindsight.
\medskip

You are shown one moment from a match, two ways:
\medskip

1. RUN-UP -- \{n\} still frames from the seconds leading up to the moment, one every \{interval\}
seconds of real time, in chronological order. Read them as a sequence: they show who is moving where,
and how fast.\\
2. LABELED FRAME -- the broadcast frame at the exact instant the decision is made. Every player is
tagged with a capital letter, and the player on the ball is marked with a gold dot. This is the
decision point.
\medskip

THE DECISION\\
The player on the ball is the one with the gold dot.
\medskip

What should he do RIGHT NOW? Your options are:\\
~~-- PASS to one of his lettered TEAMMATES. The teammates you can pass to are: \{teammates\}. Only
these letters are legal answers.\\
~~-- SHOOT.
\medskip

Reason it through first: read the shape of the defence, who is actually free, who is covered or
offside, where the space is. Ask whether a shot from here beats the best available pass. Say
concretely why your choice beats the obvious alternatives.
\medskip

Then end your reply with this line and nothing after it:
\medskip

ANSWER: <letter>
\medskip

where <letter> is one teammate letter (e.g. ANSWER: E) or the word SHOOT.
\end{promptbox}

\subsection{Belief-elicitation prompts}
\label{app:prompt-belief}
These are the prompts used to query the VLM for their estimates on the probabilities and expected values of each available action in the event (Sec.~\ref{sec:exp-belief}). Each of the two components are elicited independently, share the system message and context block below, and the model returns a ranked JSON list each time.

\begin{promptbox}[System message (decision and belief elicitation)]
You are an elite football (soccer) tactical analyst. You judge decisions the way a possession-value
model does: the best action is the one that most increases the chance your team scores in the near
future while not increasing the chance you concede. Judge the decision on what is visible at the
moment the ball is played, not on hindsight.
\end{promptbox}

\begin{promptbox}[Shared context for the two below belief queries]
You are shown one moment from a football match: a FREEZE-FRAME at the moment the on-ball player
controls the ball, and a top-down PITCH MAP of that same instant (gold star = the player on the
ball, light blue = his team T, navy = opponents O).
\medskip

Here is the map as a table. Coordinates are in metres on a 105 x 68 pitch.
\medskip

\{roster\}
\medskip

The on-ball player is marked by the gold star, at (\{actor\_x\}, \{actor\_y\}). \{attack\_sentence\}
\medskip

The player's legal actions are exactly: \{actions\}.\\
(Each letter = a pass to that teammate; SHOOT = shoot at goal.)
\end{promptbox}

\begin{promptbox}[Query 1 --- probability of success]
For each legal action, estimate the PROBABILITY IT SUCCEEDS -- a pass ``succeeds'' if it reaches the
intended teammate; a shot ``succeeds'' if it scores. Rank the actions from most likely to least
likely to succeed, and give the probability (0-1) for each.\\
Output ONLY JSON, best first: [\{"action":"A","prob":0.90\}, ...]
\end{promptbox}

\begin{promptbox}[Query 2 --- expected value]
For each legal action, estimate its EXPECTED VALUE -- how much it raises your team's chance of
scoring, accounting for BOTH how likely it is to work AND how good the result would be. Rank the
actions from best to worst expected value, and give the expected value (0-1) for each.\\
Output ONLY JSON, best first: [\{"action":"A","ev":0.30\}, ...]
\end{promptbox}

\subsection{Risk-steer prompts}
\label{app:prompt-steer}

The risk-steering experiment (Sec.~\ref{sec:exp-steer}) replaces only the deliberation paragraph of the decision prompt (App.~\ref{app:prompt-decision}, beginning “Reason it through first”) with one of the two alternatives below. The rest of the stimulus, task, and answer format are unchanged, and \textsc{neutral} is the original wording. 

\begin{promptbox}[\textsc{seek} deliberation]
Think it through first: read the shape of the defence, who is free, who is offside or covered, and
above all where the greatest attacking upside is. A safe ball is not more valuable for being secure.
Football is won by the actions that break the game open, so favour the ambitious ball --- the
penetrating pass, the runner in behind, the shot --- even when it is less likely to come off, as long
as its payoff when it succeeds justifies the chance of losing possession. Do not settle for a safe,
low-value option just because it is secure. Be concrete about why the option you pick is worth more
than the obvious safe alternatives.
\end{promptbox}

\begin{promptbox}[\textsc{avoid} deliberation]
Think it through first: read the shape of the defence, who is free, who is offside or covered, and
above all which option most reliably keeps the ball. A risky ball is not more valuable for being
ambitious. Football is lost by the cheap giveaway, so favour the secure ball --- the high-percentage
pass, the simple lay-off, the retention of possession --- even when its upside is modest, and only
take on a risky action when its payoff justifies the chance of losing possession. Do not force a
low-percentage pass or a speculative shot just because it is ambitious. Be concrete about why the
option you pick is worth more than the obvious ambitious alternatives.
\end{promptbox}

\section{Model-wise belief breakdown}
\label{app:belief-breakdown}

Table~\ref{tab:belief-recall} gives the per-model recall behind Fig.~\ref{fig:belief-sweep} at the top-33\% cutoff. Table~\ref{tab:belief-corr} gives the per-model probability--value correlation behind Fig.~\ref{fig:belief}b. 

\begin{table}[h]
\centering
\caption{Per-model recall of the best action at the top-33\% cutoff. Success-probability recall exceeds expected-value recall for most models.}
\label{tab:belief-recall}
\begin{tabular}{lcc}
\toprule
model & success-prob.\ recall & expected-value recall \\
\midrule
GPT 5.6 Sol & 79.7\% & 55.8\% \\
Gemini 3.5 Flash & 84.9\% & 50.2\% \\
Claude Opus 4.8 & 72.1\% & 58.0\% \\
\midrule
Qwen3-VL-8B & 55.8\% & 34.3\% \\
Llama-4-Maverick & 44.1\% & 59.2\% \\
\bottomrule
\end{tabular}
\end{table}

\begin{table}[h]
\centering
\caption{Per-model Spearman correlation between the model's success-probability ranking and its expected-value ranking. Every model is positive; the value model is negative.}
\label{tab:belief-corr}
\begin{tabular}{lc}
\toprule
 & $\rho$(prob, expected value) \\
\midrule
value model (ground truth) & $-0.076$ \\
\midrule
GPT 5.6 Sol & $+0.366$ \\
Gemini 3.5 Flash & $+0.371$ \\
Claude Opus 4.8 & $+0.296$ \\
Qwen3-VL-8B & $+0.521$ \\
Llama-4-Maverick & $+0.349$ \\
\bottomrule
\end{tabular}
\end{table}

\section{Value Model Robustness}
\label{app:xt}
The belief analysis in Sec.~\ref{sec:exp-belief} scores each model's estimates against VAEP. To
check that these results do not depend on that particular value model, we repeat the analysis using
Expected Threat (xT) \citep{singh2019xt}, another possession-value model that assigns each location
on the pitch a scoring probability from a move-and-shoot transition matrix. We fit xT on the same
training tournaments used for VAEP and re-value every candidate action under it. A pass is worth the
change in xT from the ball's location to the receiver's, a shot keeps its expected-goals value, and
each model's elicited beliefs are used unchanged. With the xT value model, probability of success and expected value are also nearly uncorrelated ($\rho=-0.15$), just as with VAEP. The value-recall gap persists under xT (Table~\ref{tab:xt}): the models' expected-value ranking recovers the xT-optimal action on only $32$--$38\%$ of events, if anything lower than under VAEP. The finding in Section \ref{sec:exp-belief} is therefore robust across different value models.

\begin{table}[t]
\centering
\caption{The belief results recomputed against Expected Threat (xT) in place of VAEP, pooled across both tournaments. Entries are recall@$k$ (top $33\%$): how often the model's expected-value ranking contains the value-optimal action, under each value model. The values are similar (and if anything lower) under xT, and the negative probability--value correlation reported in the text holds under both value models.}
\label{tab:xt}
\begin{tabular}{lcc}
\toprule
 & \multicolumn{2}{c}{EV recall@$k$} \\
model & VAEP & xT \\
\midrule
GPT 5.6 Sol & 56\% & 38\% \\
Gemini 3.5 Flash & 50\% & 32\% \\
Claude Opus 4.8 & 58\% & 34\% \\
\bottomrule
\end{tabular}
\end{table}

\section{The risk-aversion result is robust to value-model noise}
\label{app:robust}

The risk-aversion analysis (Sec.~\ref{sec:exp-risk}) locates each chosen action within its event's variance distribution (its risk percentile). A potential concern is that this ordering could be sensitive to noise in the estimated action values, or unstable on events where $a^\star$ only marginally exceeds the next-best action.

We alleviate this concern in two ways. First, we restrict the analysis to events in which $a^\star$ exceeds the second-best action by a clear value margin (we use a margin of 0.019, the median standard deviation among all possible actions across all events). Removing these near-tie events only sharpens the pattern: the optimal action's risk percentile rises from $76$ to $84$ and the real player's from $60$ to $70$, while every VLM stays at or below the player (between the $46$th and $64$th percentile). VLMs continue to select markedly lower-risk actions than both the optimum and the human.

Second, we perturb every action's outcome standard deviation with Gaussian noise scaled by the median action standard deviation and recompute the risk percentiles (Table~\ref{tab:robust}). The ordering is essentially invariant: the optimum stays near the $71$st--$76$th percentile and every VLM near the middle ($46$th--$53$rd), at or below the random baseline of $50$. Noise in the value model does not affect the conclusion that VLMs take on middling risk while the optimum and the professional players accept more variance.

\begin{table}[h]
\centering
\caption{Noise perturbation of the value model. Each action's outcome standard deviation is perturbed by $\mathcal{N}(0,\text{scale}\cdot\sigma_0)$ with $\sigma_0$ the median action standard deviation, and the risk percentiles are recomputed. Entries are each strategy's mean within-event risk (variance) percentile ($0$=safest, $100$=riskiest); the optimum and players stay well above the VLMs, which sit at the random baseline, at every noise level.}
\label{tab:robust}
\begin{tabular}{lcccccccc}
\toprule
noise & $a^\star$ & real player & GPT & Gemini & Claude & Qwen  & Llama & random \\
\midrule
none        & 76 & 60 & 50 & 53 & 46 & 53 & 48 & 50 \\
1$\times$sd & 72 & 59 & 51 & 52 & 47 & 52 & 48 & 50 \\
2$\times$sd & 71 & 59 & 53 & 52 & 47 & 52 & 48 & 50 \\
\bottomrule
\end{tabular}
\end{table}

\section{Labeled-frame Control}
\label{app:labeled}

To test whether performance is limited by difficulty parsing the visual input, we repeat the evaluation on the 2022 men’s World Cup subset with additional perceptual scaffolding. Models receive the same run-up video frames, but the final broadcast frame has every player tagged with a capital letter and the ball carrier explicitly marked (Figure~\ref{fig:labeled}). We omit the player map used in the default condition, allowing the labeled broadcast frame itself to provide a simpler representation of the decision context (prompt in Appendix \ref{app:prompt-labeled}). Despite this additional visual guidance, accuracy changes only modestly across models, suggesting that the observed decision-making failures are not primarily explained by difficulty identifying or localizing players.

\begin{table}[h]
\centering
\caption{Decision accuracy with player-labeled versus default visual inputs.}
\label{tab:labeled}
\begin{tabular}{lccc}
\toprule
model & labeled accuracy & default accuracy & $\Delta$ \\
\midrule
GPT 5.6 Sol       & 33.5\% & 32.3\% & $+1.1$ \\
Gemini 3.5 Flash  & 26.7\% & 27.4\% & $-0.8$ \\
Claude Opus 4.8   & 20.4\% & 19.7\% & $+0.8$ \\
\midrule
Qwen3-VL-8B       & 23.3\% & 23.1\% & $+0.2$ \\
Llama-4-Maverick  & 21.9\% & 24.6\% & $-2.6$ \\
\bottomrule
\end{tabular}
\end{table}

% \begin{figure}[h]
% \centering
% \includegraphics[width=\textwidth]{figures/good_labelled_image.png}
% \caption{Example of an image with labels for all players and the on-ball player (yellow oval) used for the easier visual control experiment (Appendix \ref{app:labeled})}
% \label{fig:labeled}
% \end{figure}

\end{document}